\documentclass[conference]{IEEEtran}
\IEEEoverridecommandlockouts

\usepackage{cite}
\usepackage{amsmath,amssymb,amsfonts}
\usepackage{algorithmic}
\usepackage{graphicx}
\usepackage{textcomp}
\usepackage{xcolor}
\usepackage{hyperref}
\usepackage{booktabs}
\def\BibTeX{{\rm B\kern-.05em{\sc i\kern-.025em b}\kern-.08em
    T\kern-.1667em\lower.7ex\hbox{E}\kern-.125emX}}
\begin{document}

\title{MosquitoMiner: A Light Weight Rover for Detecting and Eliminating Mosquito Breeding Sites
}

\author{
\IEEEauthorblockN{1\textsuperscript{st} Md. Adnanul Islam}
\IEEEauthorblockA{\textit{Computer Science and Engineering} \\
\textit{United International University}\\
Dhaka, Bangladesh \\
mislam221096@bscse.uiu.ac.bd}
\and
\IEEEauthorblockN{2\textsuperscript{nd} Md. Faiyaz Abdullah Sayeedi}
\IEEEauthorblockA{\textit{Computer Science and Engineering} \\
\textit{United International University}\\
Dhaka, Bangladesh \\
msayeedi212049@bscse.uiu.ac.bd}
\and
\IEEEauthorblockN{3\textsuperscript{rd} Jannatul Ferdous Deepti}
\IEEEauthorblockA{\textit{Computer Science and Engineering} \\
\textit{United International University}\\
Dhaka, Bangladesh \\
jdeepti212008@bscse.uiu.ac.bd }
\and
\IEEEauthorblockN{4\textsuperscript{th} Shahanur Rahman Bappy}
\IEEEauthorblockA{\textit{Computer Science and Engineering} \\
\textit{United International University}\\
Dhaka, Bangladesh \\
sbappy211002@bscse.uiu.ac.bd}
\and
\IEEEauthorblockN{5\textsuperscript{th} Safrin Sanzida Islam}
\IEEEauthorblockA{\textit{Computer Science and Engineering} \\
\textit{United International University}\\
Dhaka, Bangladesh \\
sislam212124@bscse.uiu.ac.bd }
\and
\IEEEauthorblockN{6\textsuperscript{th} Fahim Hafiz}
\IEEEauthorblockA{\textit{Computer Science and Engineering} \\
\textit{United International University}\\
Dhaka, Bangladesh\\
fahimhafiz@cse.uiu.ac.bd}
}

\maketitle

\begin{abstract}
In this paper, we present a novel approach to the development and deployment of an autonomous mosquito breeding place detector rover with the object and obstacle detection capabilities to control mosquitoes. Mosquito-borne diseases continue to pose significant health threats globally, with conventional control methods proving slow and inefficient. Amidst rising concerns over the rapid spread of these diseases, there is an urgent need for innovative and efficient strategies to manage mosquito populations and prevent disease transmission. To mitigate the limitations of manual labor and traditional methods, our rover employs autonomous control strategies. Leveraging our own custom dataset, the rover can autonomously navigate along a pre-defined path, identifying and mitigating potential breeding grounds with precision. It then proceeds to eliminate these breeding grounds by spraying a chemical agent, effectively eradicating mosquito habitats. Our project demonstrates the effectiveness that is absent in traditional ways of controlling and safeguarding public health. The code for this project is available on GitHub at - \url{https://github.com/faiyazabdullah/MosquitoMiner}
\end{abstract}


\section{Introduction}
Nowadays the global battle against mosquito-transmitted diseases has intensified. Mosquito-born diseases like malaria, dengue fever, zika virus disease, and others continue to pose significant threats to public health\cite{mosquito_borne_disease}. People are facing difficulties in controlling the rise of mosquitoes and their impact on health. As mosquitoes are getting stronger against bug-killing sprays, that's why people are facing problems while controlling them, we need innovations to tackle the problem.

Motivated by the needs of society, our team jumped into a mission to develop a sophisticated mosquito-breeding place detection rover. It's quite hard to stop mosquitoes from spreading diseases, but finding and stopping mosquitoes in their breeding place is mostly important to stop their journey of spreading disease. On the other hand, manually checking out their breeding place one by one is time-consuming and less effective. So, we are eager to use technology to do this work as perfectly as possible \cite{hotspot_detection_dengue_auto_robotics}.

Our project focuses on building an automatic or self-driving rover that can easily find mosquito breeding spots. We have made our custom local dataset \cite{mosquito_faiyaz} to feed the machine learning model so that the rover can understand the pattern of mosquito hotspots \cite{insect_detection_rover}. Our rover can go along a pre-selected path and check for places where mosquitoes might breed in common, based on the dataset it has trained. If it finds any mosquito breeding place then it will use bug-killing spray and go forward. After reaching the ending point it will automatically come back to the starting point it has started. In terms of any issues, we can also manually handle this rover.

Our project has had significant importance because it could change the mechanism of controlling mosquitoes and keep public health secure from mosquito-borne disease \cite{tecnology_for_control}. We are finding out the hotspots of mosquitoes as early as possible, which can eliminate the illness before it spreads in the environment.

The rest of this paper is structured as follows: Section \ref{lir_review} covers the literature review. Section \ref{system} describes the system design. Section \ref{software} discusses the software implementation. Section \ref{eval_matrices} outlines the evaluation metrics. Section \ref{result} presents the result analysis. Section \ref{conclusion} provides the conclusion.

\section{Literature Review}
\label{lir_review}
Innovation has become a necessity to surveillance and control the mosquito-born disease that poses significant public health challenges worldwide. In this literature review, we have gone through the existing research landscapes about mosquito control strategies, with a focus on the limitations in traditional methods and the way advancements in autonomous rover systems \cite{rover_design} \cite{insect_detection_rover}.

As we know traditional mosquito control methods mainly depend on manual inspection and insecticide application, which is time-consuming, labor-intensive, most importantly ineffective in controlling the mosquitoes. Many studies have pointed out the inefficiency and challenges of traditional approaches, emphasizing the need for alternative solutions \cite{tecnology_for_control}.

Recently, developments in automation technology have paved the way for new applications for mosquito control \cite{rover_automatic} \cite{rover_ground_object_detection}. Equipped with advanced features such as sensors, complex processors, robust AI, and ML models, the automation platform of mosquito control strategies can be modified by rapid detection of mosquito breeding sites with perceived accuracy \cite{autonomous_rover_object_detection}.

On the other hand, a lot of studies have been done on automation on rovers \cite{autonomous_rover}. This automation is nowadays so advanced that it has been proposed as an important solution for mosquito surveillance. These rovers are attached with high-quality cameras, sensors, and processors with detection capacity \cite{object_detection_rover}. Many of the rover's structures have been designed based on the Mars Rover, ensuring it can traverse rough terrain smoothly.

Furthermore, machine learning models have also become advanced from time to time \cite{ml_for_breeding_place}. Machine learning and computer vision have enabled the path of object detection. By leveraging a large dataset of mosquitoes, researchers have trained many models for the detection of mosquitoes and a lot of work has been done about mosquito control as well but there is no significant result or outcome for their research \cite{ml_for_breeding_place}. They were talking about alternative approaches but could not implement this type of thing to control mosquitoes.

Ultimately, challenges remain in developing a fully functional model for societal well-being through automation and detection for mosquito-borne disease control \cite{ml_for_mosquito_control}. To enhance deployment, ensuring robustness is essential. It is crucial to establish a safe and efficient system for monitoring and controlling autonomous vehicles to advance technology \cite{rover_ground_object_detection}.

Overall, the literature suggests that, in this era, automation holds a promising opportunity for improving mosquito surveillance and control. By leveraging the capabilities of robotics and artificial intelligence, we can have the potential to develop more effective strategies for mosquito-controlling missions to protect public health by stopping the spreading of mosquito-borne diseases.

\section{System Overview}
\label{system}
Figure \ref{Fig2_sys_arch} shows the overall overview of the system architecture.

\subsection{Hardware Components}

\begin{figure}[h]
  \centering
  \includegraphics[height=5cm, width=\linewidth]{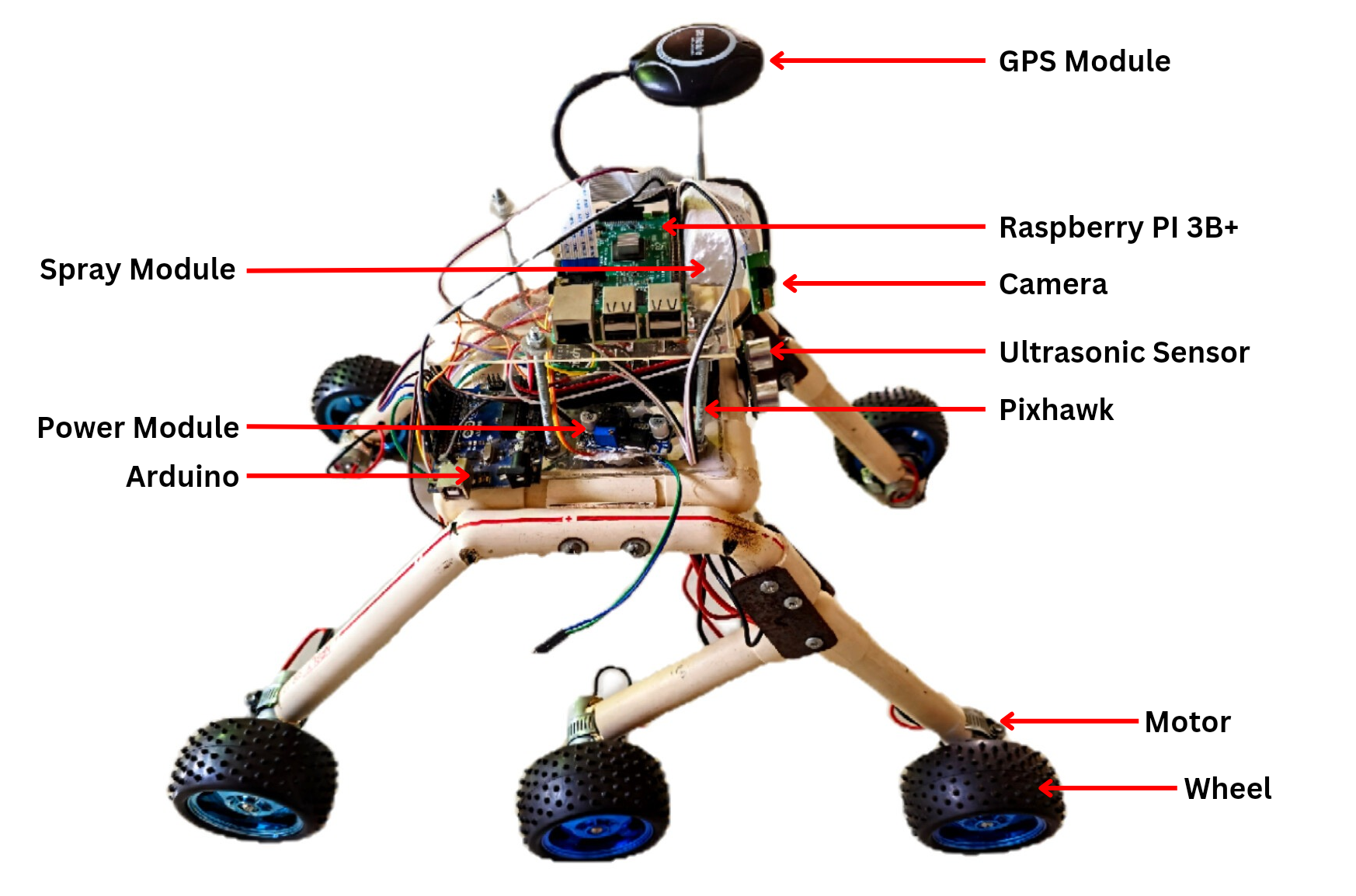}
  \caption{Exploded View of the Hardware Components}
  \label{fig1_exploded_view}
\end{figure}

\begin{figure}[h]
  \centering
  \includegraphics[width=\linewidth]{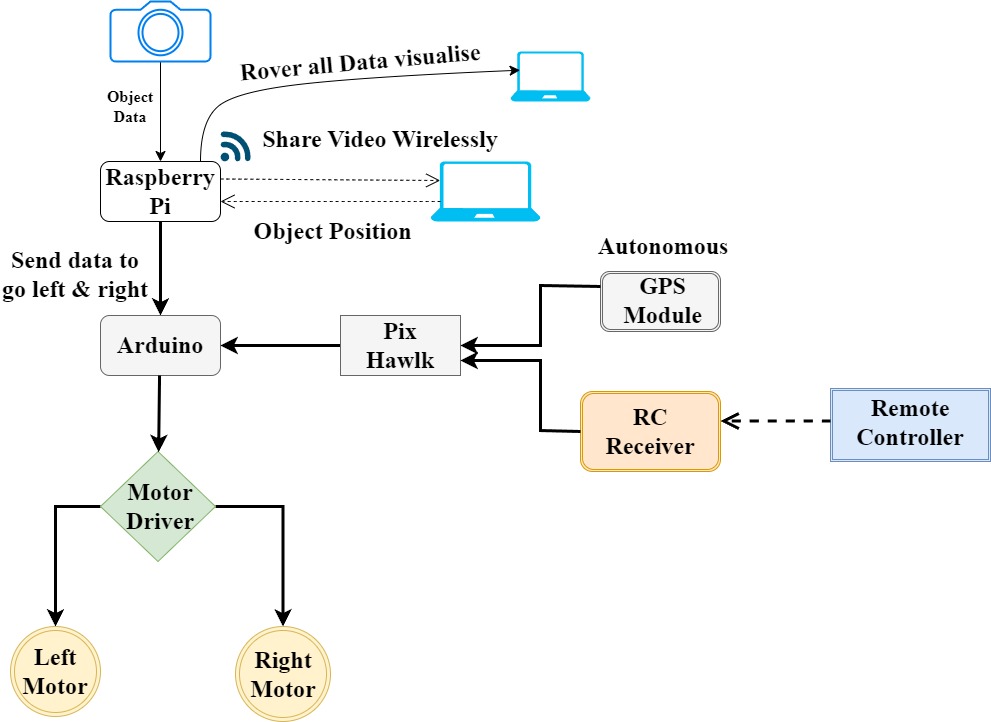}
  \caption{Overview of our System Architecture}
  \label{Fig2_sys_arch}
\end{figure}

The exploded view of our rover is shown in Figure \ref{fig1_exploded_view}. The detailed description of the hardware components we used are described below:

\subsubsection{Raspberry Pi 3 Model B}
The central processing unit for the MosquitoMiner. It controls movement, processes sensor data, and executes mosquito breeding site detection and elimination algorithms.
\subsubsection{Raspberry Pi Camera Module}
A camera mounted on the rover captures visual data of the surrounding environment. This data is used by the system to identify potential mosquito breeding sites.
\subsubsection{GPS Module}
The GPS module provides the rover with its location data. This information is crucial for navigation, tracking the rover's movements, and potentially creating maps of mosquito breeding site locations.
\subsubsection{Arduino Uno}
This microcontroller board is responsible for lower-level tasks like sensor data acquisition (e.g., from ultrasonic sensors for obstacle detection), motor control communication with the motor driver, or interfacing with other auxiliary components.
\subsubsection{Pixhawk (or Flight Controller)}
This autopilot system offers advanced functionalities for autonomous navigation, including flight stabilization, path planning, and waypoint control. It is used in conjunction with the Raspberry Pi for more complex movement and obstacle avoidance maneuvers.
\subsubsection{Remote Controller}
The remote controller allows a human operator to manually control the rover's movements. This might be useful for maneuvering the rover in complex terrain or during initial setup and testing.
\subsubsection{Motor Driver}
The motor driver circuit controls the rover's motors, enabling it to move forward, backward, turn, and potentially adjust its speed.
\subsubsection{Power Module}
Supplies and manages power distribution to all electronic components on the rover, ensuring stable operation and system safety.
\subsubsection{Ultrasonic Sensor}
Used for obstacle detection. These sensors help the rover navigate by detecting objects in its path and determining the distance to them, aiding in collision avoidance.
\subsubsection{Spray module}
The spray module integrated into the MosquitoMiner rover is designed to efficiently dispense larvicides at identified mosquito breeding sites, ensuring targeted and effective elimination of mosquito larvae.

\subsection{Data Acquisition}
The MosquitoMiner's journey starts with its camera capturing visual data of the surrounding environment. This data is transmitted to the Raspberry Pi, the central processing unit. The GPS module continuously acquires the rover's location, sending this data to the Raspberry Pi. This information is crucial for navigation, tracking movements, and potentially creating maps of breeding sites. Arduino microcontroller collects data from additional sensors like ultrasonic sensors for obstacle detection or other environmental sensors relevant to mosquito breeding. This sensor data is then sent to the Raspberry Pi for further processing.

\subsection{Data Processing and Decision Making}
The Raspberry Pi receives the camera data and utilizes image recognition algorithms to identify potential mosquito breeding sites within the captured visuals. It also integrates the received GPS data into a navigation system or uses it for data logging purposes. The Raspberry Pi performs any necessary pre-processing or filtering on the sensor data it receives. Based on this processed data (identified breeding sites, location, and potential obstacles), the Raspberry Pi makes critical decisions about the rover's actions. The Raspberry Pi communicates with the Pixhawk to share information about detected breeding sites and planned paths for optimal navigation.

The Pixhawk flight controller receives information about detected breeding sites and planned paths from the Raspberry Pi. It utilizes its own sensors (gyroscope, accelerometer) for flight stabilization and integrates this data with the Raspberry Pi information for more complex navigation decisions. The Pixhawk utilizes path-planning algorithms to determine optimal routes between the rover's current location and identified breeding sites, considering potential obstacles based on sensor data.

\subsection{Movement Control and Elimination}
Based on the decision-making process, control signals are transmitted from the central processing unit (Raspberry Pi) to the motor driver. These control signals dictate the direction and speed of the rover's movement.
The motor driver receives control signals from the Arduino. It regulates the left and right motors based on these signals, causing the rover to move forward, backward, turn, or adjust speed to navigate towards or away from breeding sites while potentially avoiding obstacles.
Upon reaching a breeding site, the MosquitoMiner employs a mechanism to eliminate or disrupt the breeding ground. It sprays a larvicide using a dispenser (Spray Module) controlled by the Arduino.

\subsection{Human Intervention}
The operator can take manual control of the rover's movement using the remote controller, overriding any autonomous decision-making and allowing for manual navigation or adjustments.

\section{Software Architecture}
\label{software}
This section describes the software architecture of our rover system designed to detect and eliminate mosquito breeding sites. Figure \ref{fig:soft} shows the overview of our software architecture.

\subsection{Software Components}

\subsubsection{Operating System (OS)}
The Raspberry Pi likely runs a lightweight operating system such as Raspbian OS, which provides a platform for executing the software components and managing hardware resources.

\subsubsection{MAVProxy (Ground Control Station)}
MAVProxy acts as the ground control station software, providing a user interface for mission planning, telemetry display, and manual control of the rover. It communicates with the rover using a wireless connection.

\subsubsection{MissionPlanner}
MissionPlanner is used as an alternative ground control station software, offering similar functionalities to MAVProxy for mission planning, telemetry viewing, and manual control.

\begin{figure}[h]
  \centering
  \includegraphics[width=\linewidth]{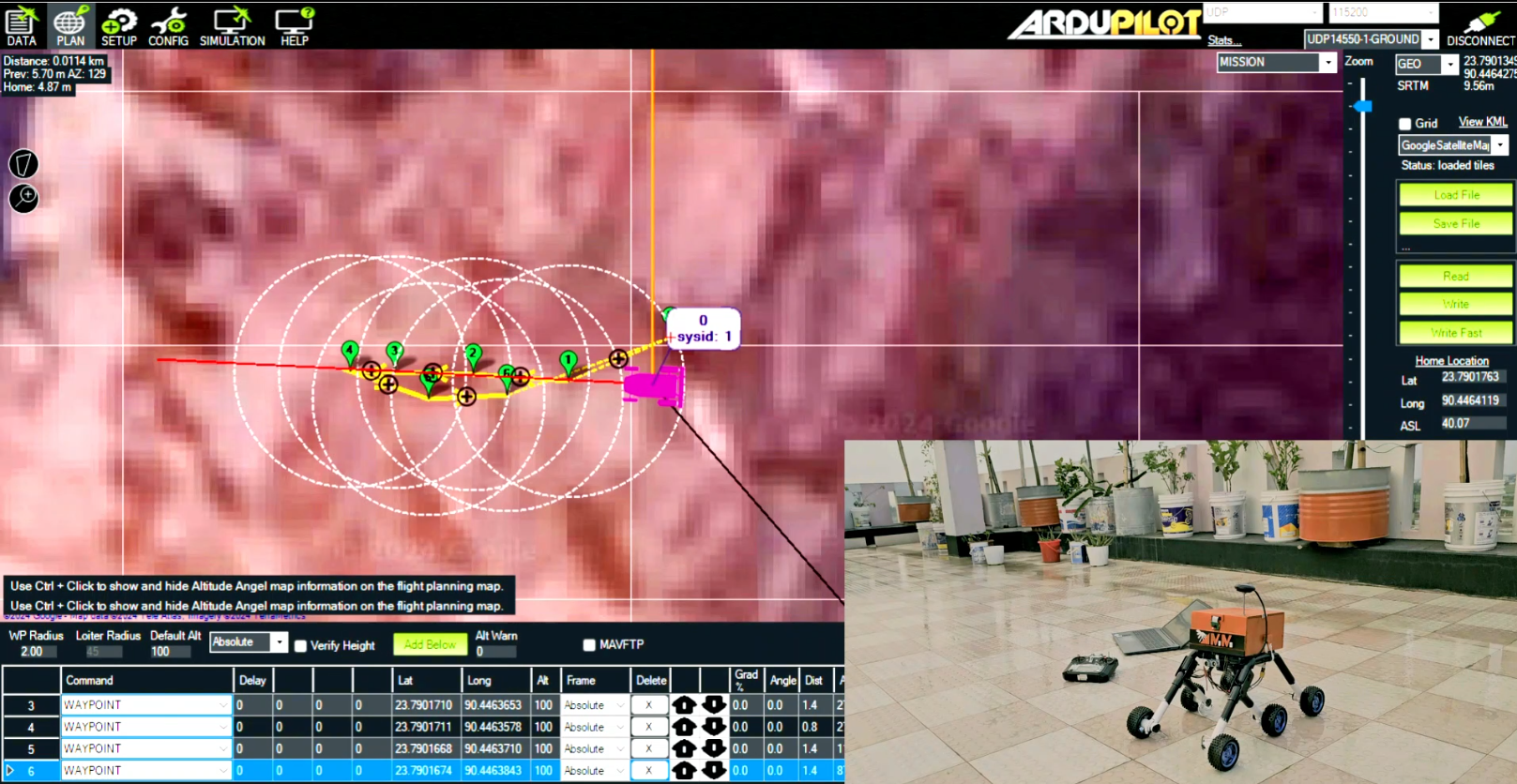}
  \caption{Calibration of MissionPlanner Ardupilot with MosquitoMiner}
\end{figure}

\subsubsection{OpenCV}
OpenCV (Open Source Computer Vision Library) is a software library providing image and video processing capabilities. It is used for tasks like camera calibration and image pre-processing before object detection.

\subsubsection{YOLOv8s (You Only Look Once)}
YOLOv8s is a convolutional neutral network (CNN) based deep learning model for object detection. The Raspberry Pi utilizes YOLOv8 to identify potential mosquito breeding sites within the captured camera frames.

\begin{figure}[h]
  \centering
  \includegraphics[height=5cm, width=\linewidth]{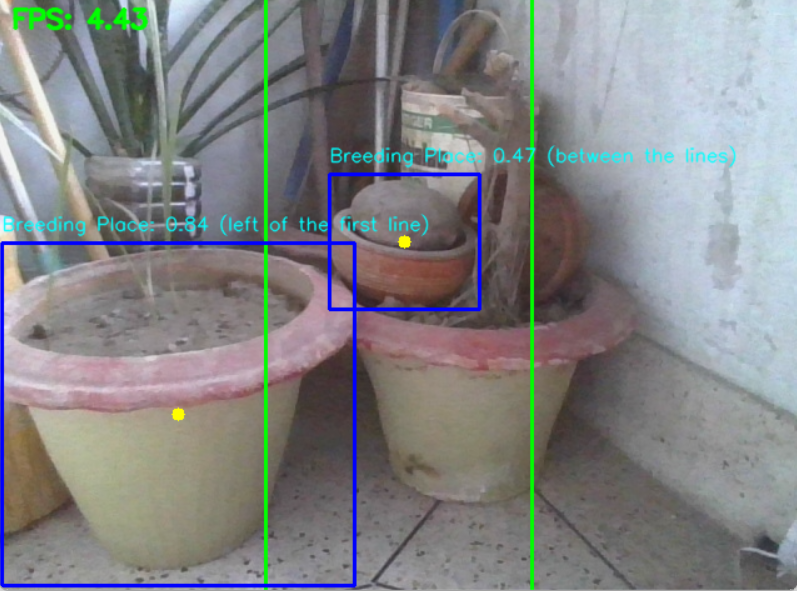}
  \caption{Real-time object detection and position determination, with the green lines indicating the center of the rover.}
\end{figure}

\subsection{Data Flow and Communication}

The Raspberry Pi camera captures video or image data of the surrounding environment. Then the OpenCV library is used for image pre-processing tasks (e.g., resizing, and color adjustments) before feeding the data to the object detection model. YOLOv8s model analyzes the pre-processed image data to detect and classify potential mosquito breeding sites within the frames. MAVProxy or MissionPlanner receives telemetry data from the rover, including sensor readings, GPS location, and potential object detection results for visualization or logging purposes. The user interacts with MAVProxy or MissionPlanner to send control commands to the rover, such as navigation waypoints or manual control inputs. These commands are transmitted wirelessly to the rover.

\begin{figure}[h]
  \centering
  \includegraphics[width=\linewidth]{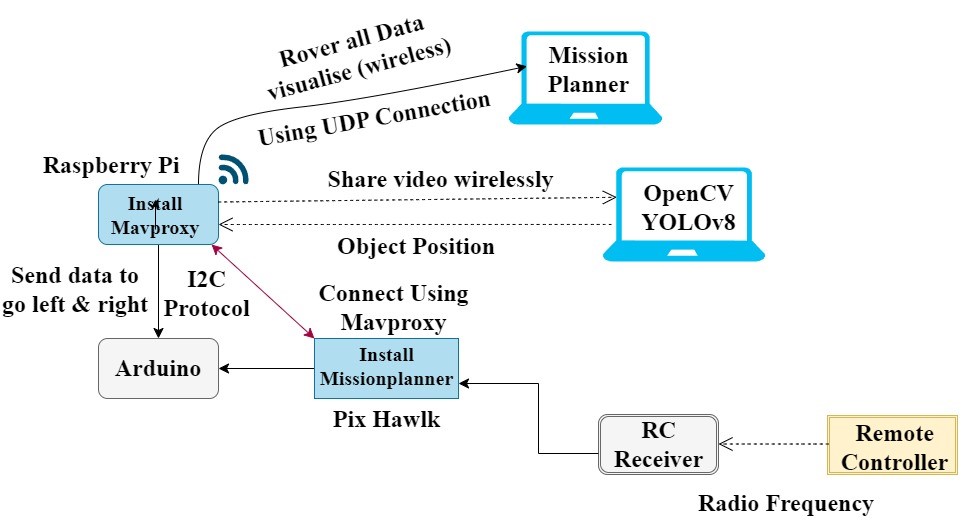}
  \caption{Overview of our Software Architecture}
  \label{fig:soft}
\end{figure}

\subsection{Rover Control}
The Raspberry Pi receives control signals from the ground control station software (MAVProxy or MissionPlanner). It utilizes a motor control library to interpret the received commands and translate them into signals for the rover's motors, enabling movement control.

\subsection{Communication Protocols}
This project utilized wireless communication - User Datagram Protocol (UDP) between the rover Raspberry Pi and the ground control station software - MAVProxy and MissionPlanner.

\section{Evaluation Matrices}
\label{eval_matrices}
\subsection{Detection Accuracy}

\subsubsection{Mean Average Precision (mAP@50)} This metric calculates the average precision at an Intersection Over Union (IoU) threshold of 50\%. It is the mean of the Average Precision (AP) for each class at this threshold.
\begin{equation}
    \text{mAP@50} = \frac{1}{N} \sum_{i=1}^{N} AP_i
\end{equation}
where, \(N\) is the number of classes (in this case, mosquito and breeding sites).
    
\subsubsection{Precision} The ratio of correctly predicted positive observations to the total predicted positives. It helps to measure the accuracy of the detections.
\begin{equation}
    \text{Precision} = \frac{TP}{TP + FP}
\end{equation}
where, \(TP\) and \(FP\) stand for True Positives and False Positives, respectively.
    
\subsubsection{Recall} The ratio of correctly predicted positive observations to all observations in actual class. It measures the ability of the model to find all the relevant cases.
\begin{equation}
    \text{Recall} = \frac{TP}{TP + FN}
\end{equation}
where, \(FN\) is the number of False Negatives.

\subsection{Elimination Effectiveness}

\subsubsection{Target Class Reduction Rate} The percentage reduction in the population of the target class (mosquito and breeding sites) post-treatment compared to pre-treatment levels.
\begin{equation}
    \text{TCRR} = \left(1 - \frac{P_{\text{post}}}{P_{\text{pre}}}\right) \times 100\%
\end{equation}

where, \( P_{\text{post}} \) represents the post-treatment population and \( P_{\text{pre}} \) represents the pre-treatment population.
    
\subsubsection{Area Coverage} The percentage of the total designated area that the rover can survey and treat effectively.
\begin{equation}
    \text{Area Coverage} = \left(\frac{\text{Area Treated}}{\text{Total Area}}\right) \times 100\%
\end{equation}

\subsection{Operational Efficiency}

\subsubsection{Battery Life} The total operational time the rover can run on a single charge or before needing a recharge.
\subsubsection{Time to Complete a Mission} The total time taken by the rover to complete a survey and treatment mission, is measured in minutes.
\subsubsection{Cost Effectiveness} The overall affordability and economic efficiency of deploying and operating the rover.

\section{Result Analysis}
\label{result}
\subsection{Detection Accuracy}
The performance of the MosquitoMiner in detecting mosquitoes and breeding sites has been quantitatively assessed. The system used YOLOv8s pre-trained model and fine-tuned on our custom dataset, achieving a Mean Average Precision (mAP@50) of 61.7\%, indicating a moderate level of accuracy in detecting relevant objects within a 50\% intersection over union threshold. The precision of the system, which measures the accuracy of positive predictions, was recorded at 84.7\%. The recall rate, which measures the system’s ability to detect all relevant instances, was observed at 51.6\%. These figures suggest that while the system is relatively reliable in its predictions, there remains room for improvement in detecting all potential threats. The summary of detection accuracy metrics is shown in Table \ref{tab:detection_accuracy}.

\begin{table}[h]
\centering
\caption{Summary of Detection Accuracy Metrics}
\label{tab:detection_accuracy}
\begin{tabular}{@{}lc@{}}
\toprule
Metric                     & Value (\%) \\ \midrule
Mean Average Precision (mAP@50) & 61.7     \\
Precision                  & 84.7      \\
Recall                     & 51.6      \\ \bottomrule
\end{tabular}
\end{table}

\begin{figure}[h]
  \centering
  \includegraphics[height=5cm, width=\linewidth]{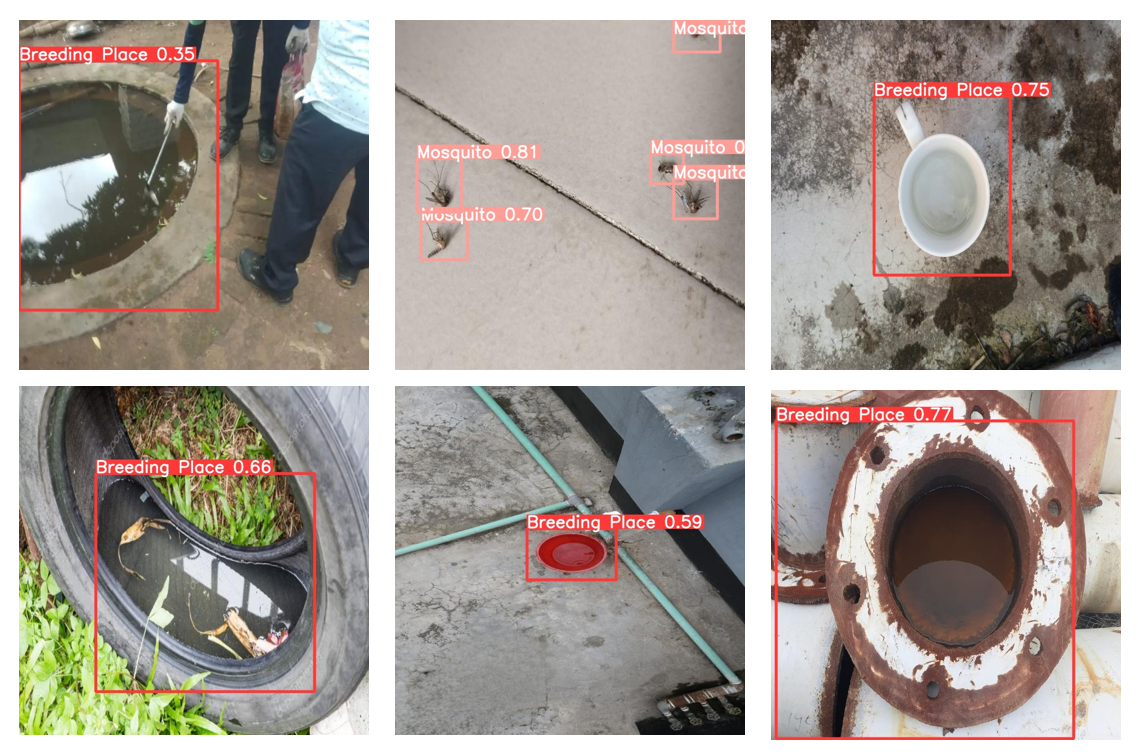}
  \caption{Various Environmental Backgrounds for More Real-time Detection}
\end{figure}

\subsection{Elimination Effectiveness}

\subsubsection{Target Class Reduction Rate}
To quantitatively assess the effectiveness of our autonomous MosquitoMiner rover in reducing mosquito populations in targeted breeding sites, we employed a specific metric known as the Target Class Reduction Rate (TCRR).

For this study, we designed and constructed a controlled experimental environment that mimicked real-world conditions where mosquitoes typically breed. The setup included five distinct artificial breeding sites, each created to represent common mosquito habitats. These sites were strategically placed within a designated test area and were uniformly distributed to ensure that each site was accessible by the rover under test conditions. The rover was programmed to autonomously navigate through the test area, detect breeding sites using its integrated camera, and then apply a targeted treatment designed to eliminate the breeding potential at each site.

The rover successfully identified and reduced the viability of four out of the five breeding sites. These results underscore the efficiency of the MosquitoMiner in autonomously navigating and treating mosquito breeding sites, thereby reducing the number of active breeding grounds by 80\%. This significant reduction in breeding sites indicates a potential for a corresponding decrease in the mosquito population, which is pivotal for controlling the spread of mosquito-borne diseases.

\subsubsection{Area Coverage}
Our study also examined the navigational accuracy and area coverage capabilities of the MosquitoMiner rover. To do this, we incorporated a predetermined path within our experimental setup, marked by a series of eight nodes that served as checkpoints representing significant waypoints within the target environment. The predefined path was carefully mapped to ensure comprehensive coverage of the area and included various terrains and obstacles typical of mosquito-prone environments. Each of the eight nodes was strategically placed to guide the rover through critical areas, maximizing the exposure to all potential breeding sites.

The MosquitoMiner was programmed to autonomously navigate this path using its integrated GPS and obstacle detection systems. The success of the navigation was measured by the rover's ability to reach these checkpoints without manual intervention.

\begin{figure}[h]
  \centering
  \includegraphics[width=\linewidth]{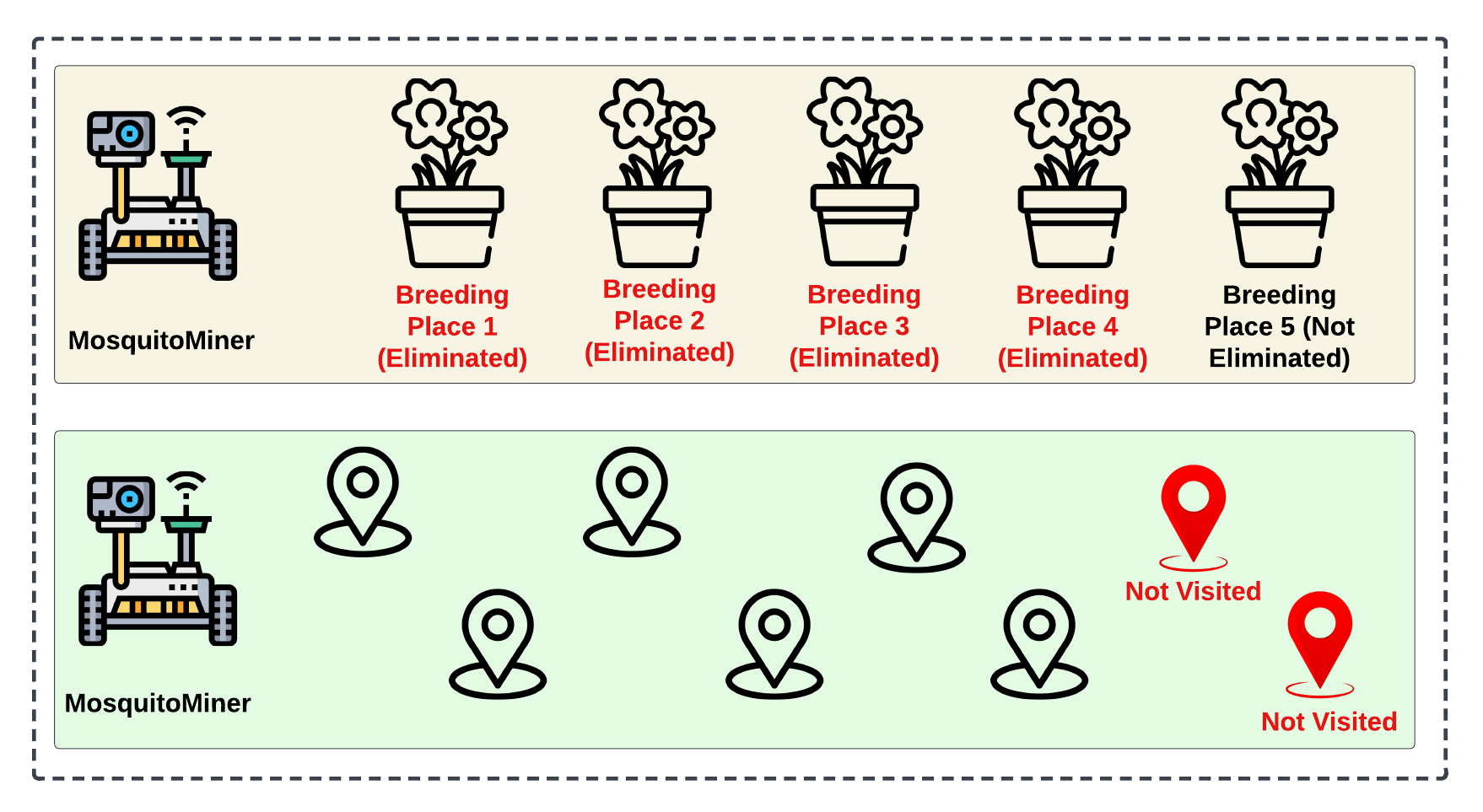}
  \caption{Overview of the Experimental Setup for \textbf{(Up)} Target Class Reduction Rate (80\%) \textbf{(Down)} Area Coverage (75\%)}
  \label{fig:exp_set}
\end{figure}

The rover successfully reached six out of the eight designated checkpoints, translating to a 75\% success rate in area coverage. This indicates that the rover effectively covered 75\% of the designated area, ensuring that the majority of the target environment was treated. Figure \ref{fig:exp_set} shows the overview of the experimental setup.



\begin{table}[h]
\centering
\caption{Total Cost Calculation for the MosquitoMiner Rover}
\label{tab:cost}
\begin{tabular}{@{}lc@{}}
\toprule
Component         & Price in USD (\$)       \\ \midrule
Motor             & 6 * 7.27 = 43.62 \\
Wheel             & 6 * 3.85 = 23.10 \\
Hex Coupling      & 6 * 0.94 = 5.64  \\
Pipe              & 10.26            \\
GPS Module        & 19.66            \\
Wire              & 4.27             \\
Nut               & 2.56             \\
Pixhawk           & 111.12           \\
Motor Driver      & 8.55             \\
Camera            & 6.84             \\
Raspberry Pi 3B   & 85.47            \\
Arduino           & 3.85             \\
Remote Controller & 59.83            \\
Battery 3000mAh   & 25.64            \\
\midrule
Total             & 409.39           \\ \bottomrule
\end{tabular}
\end{table}

\subsection{Operational Efficiency}
Operational efficiency metrics provide insight into the practical deployment of the rover. The battery life of the system is robust, allowing for extended operational periods before requiring a recharge. This feature enhances the rover's usability in field conditions. Additionally, the cost-effectiveness of the MosquitoMiner is notable, with the total operational cost being around 410 USD shown in Table \ref{tab:cost}, making it more affordable compared to other solutions in the market \cite{zhang2023designing} \cite{srisuphab2018insect}. This cost efficiency, combined with operational effectiveness, makes the MosquitoMiner a valuable tool in combating mosquito-borne diseases. The time to complete the mission was meticulously measured from the moment the rover was activated to when it reached the final checkpoint on the predefined path. In our controlled test environment, the MosquitoMiner completed its mission-navigating through the predetermined path and executing its detection and elimination tasks in just 9 minutes and 43 seconds. This performance indicator is particularly significant, illustrating the rover's potential for quick deployment and operation. 

\section{Conclusion}
\label{conclusion}
\begin{figure}[h]
  \centering
  \includegraphics[height=5.5cm, width=\linewidth]{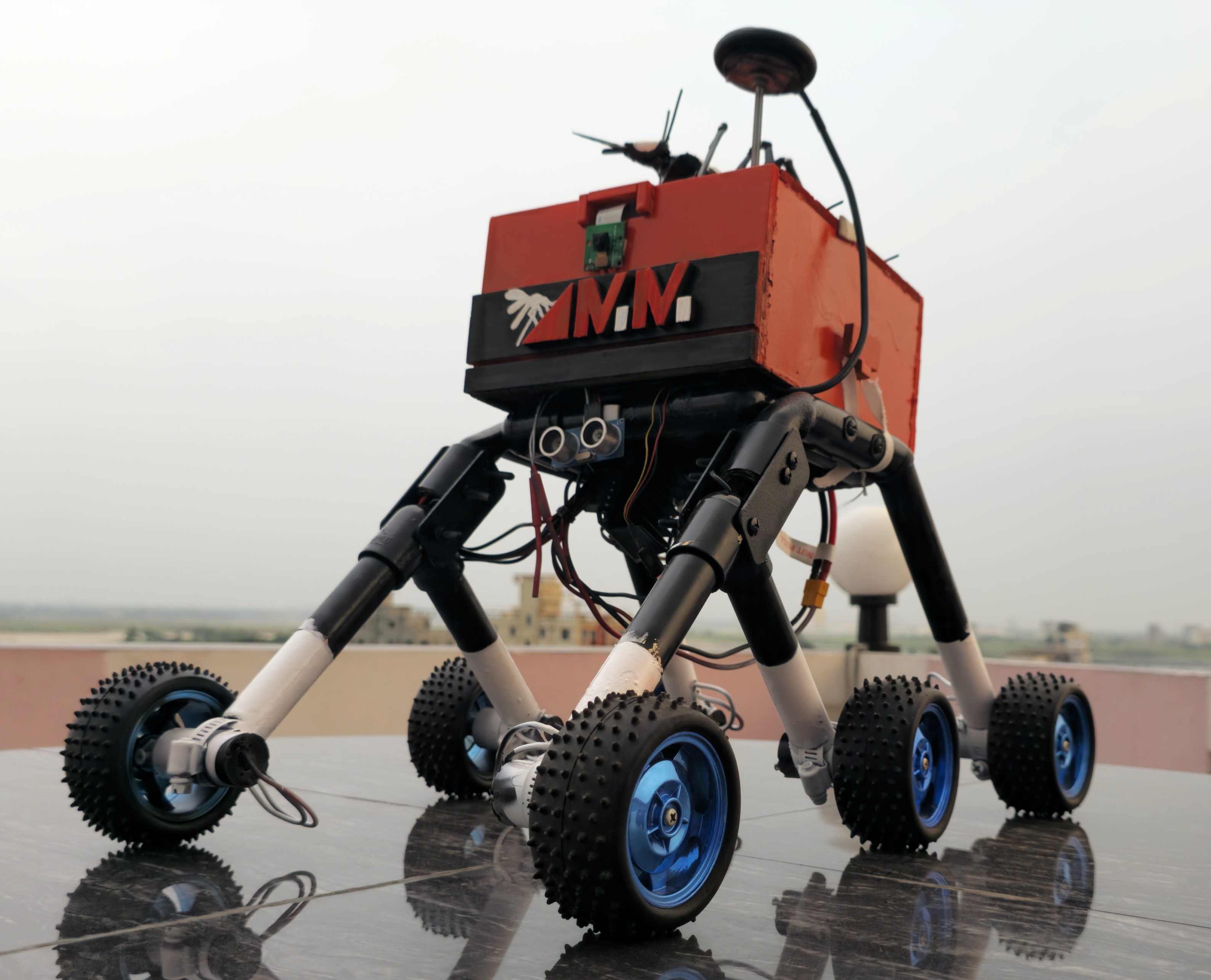}
  \caption{The Implemented MosquitoMiner Rover}
  \label{fig:rover_image}
\end{figure}

Our project, shown in Figure \ref{fig:rover_image}, introduces a promising solution for controlling mosquitoes through the development of our autonomous breeding place detector rover. By using advanced techniques and autonomous control strategies, we have already demonstrated the efficiency of our implementation in mitigating traditional limitations like manual labor methods for safeguarding public health from mosquito-borne disease. Our future work will be upgrading our rover's present hardware, especially for smoothness, and will replace Raspberry Pi 3 with a more sophisticated microcontroller, which will enhance the rover's processing power, efficiency, and robust performance. Moreover, we will enhance our dataset with more compatible data so that our rover can understand different patterns of breeding places. Also, we will enhance our rover structurally by integrating other components for detailing and allowing it to navigate and operate effectively in challenging environments.

\bibliographystyle{IEEEtran}
\bibliography{reference.bib}

\begin{thebibliography}{10}
\providecommand{\url}[1]{#1}
\csname url@samestyle\endcsname
\providecommand{\newblock}{\relax}
\providecommand{\bibinfo}[2]{#2}
\providecommand{\BIBentrySTDinterwordspacing}{\spaceskip=0pt\relax}
\providecommand{\BIBentryALTinterwordstretchfactor}{4}
\providecommand{\BIBentryALTinterwordspacing}{\spaceskip=\fontdimen2\font plus
\BIBentryALTinterwordstretchfactor\fontdimen3\font minus \fontdimen4\font\relax}
\providecommand{\BIBforeignlanguage}[2]{{%
\expandafter\ifx\csname l@#1\endcsname\relax
\typeout{** WARNING: IEEEtran.bst: No hyphenation pattern has been}%
\typeout{** loaded for the language `#1'. Using the pattern for}%
\typeout{** the default language instead.}%
\else
\language=\csname l@#1\endcsname
\fi
#2}}
\providecommand{\BIBdecl}{\relax}
\BIBdecl

\bibitem{mosquito_borne_disease}
M.~A. Tolle, ``Mosquito-borne diseases,'' \emph{Current problems in pediatric and adolescent health care}, vol.~39, no.~4, pp. 97--140, 2009.

\bibitem{hotspot_detection_dengue_auto_robotics}
S.~Sreeram and L.~Shanmugam, ``Autonomous robotic system based environmental assessment and dengue hot-spot identification,'' in \emph{2018 IEEE International Conference on Environment and Electrical Engineering and 2018 IEEE Industrial and Commercial Power Systems Europe (EEEIC/I\&CPS Europe)}.\hskip 1em plus 0.5em minus 0.4em\relax IEEE, 2018, pp. 1--6.

\bibitem{mosquito_faiyaz}
\BIBentryALTinterwordspacing
M.~F.~A. Sayeedi, F.~Hafiz, and M.~A. Rahman, ``Mosquitofusion: A multiclass dataset for real-time detection of mosquitoes, swarms, and breeding sites using deep learning,'' in \emph{The Second Tiny Papers Track at ICLR 2024}, 2024. [Online]. Available: \url{https://openreview.net/forum?id=3s4hFx8pYs}
\BIBentrySTDinterwordspacing

\bibitem{insect_detection_rover}
A.~Srisuphab, P.~Silapachote, W.~Tantratorn, P.~Krakornkul, and P.~Darote, ``Insect detection on an unmanned ground rover,'' in \emph{TENCON 2018-2018 IEEE Region 10 Conference}.\hskip 1em plus 0.5em minus 0.4em\relax IEEE, 2018, pp. 0954--0959.

\bibitem{tecnology_for_control}
E.~J. Norris and J.~R. Coats, ``Current and future repellent technologies: the potential of spatial repellents and their place in mosquito-borne disease control,'' \emph{International journal of environmental research and public health}, vol.~14, no.~2, p. 124, 2017.

\bibitem{rover_design}
S.~Aswath, V.~Unnikrishnan, T.~K. Sreekuttan, H.~Abin~Simon, P.~M. Deepu, M.~Menon, P.~Basil, R.~Ramachandran, A.~Sengar, A.~Balakrishnan \emph{et~al.}, ``Design and development of an intelligent rover for mars exploration,'' in \emph{The 18th Annual International Mars Society Convention}, 2016.

\bibitem{rover_automatic}
A.~Howard and H.~Seraji, ``A real-time autonomous rover navigation system,'' in \emph{Proceedings from World Automation Congress (2000)}.\hskip 1em plus 0.5em minus 0.4em\relax Citeseer, 2000.

\bibitem{rover_ground_object_detection}
S.~H. Hussain, A.~Hussain, R.~H. Shah, and S.~A. Abro, ``Mini rover-object detecting ground vehicle (ugv),'' \emph{University of Sindh Journal of Information and Communication Technology}, vol.~3, no.~2, pp. 104--108, 2019.

\bibitem{autonomous_rover_object_detection}
Y.~Chen, X.~Chen, J.~Zhu, F.~Lin, and B.~M. Chen, ``Development of an autonomous unmanned surface vehicle with object detection using deep learning,'' in \emph{IECON 2018-44th annual conference of the IEEE industrial electronics society}.\hskip 1em plus 0.5em minus 0.4em\relax IEEE, 2018, pp. 5636--5641.

\bibitem{autonomous_rover}
M.~J. Schuster, C.~Brand, S.~G. Brunner, P.~Lehner, J.~Reill, S.~Riedel, T.~Bodenm{\"u}ller, K.~Bussmann, S.~B{\"u}ttner, A.~D{\"o}mel \emph{et~al.}, ``The lru rover for autonomous planetary exploration and its success in the spacebotcamp challenge,'' in \emph{2016 International Conference on Autonomous Robot Systems and Competitions (ICARSC)}.\hskip 1em plus 0.5em minus 0.4em\relax IEEE, 2016, pp. 7--14.

\bibitem{object_detection_rover}
J.~H. Lim and S.~K. Phang, ``Classification and detection of obstacles for rover navigation,'' in \emph{Journal of Physics: Conference Series}, vol. 2523, no.~1.\hskip 1em plus 0.5em minus 0.4em\relax IOP Publishing, 2023, p. 012030.

\bibitem{ml_for_breeding_place}
D.~T. Bravo, G.~A. Lima, W.~A.~L. Alves, V.~P. Colombo, L.~Djogbenou, S.~V.~D. Pamboukian, C.~C. Quaresma, and S.~A. de~Araujo, ``Automatic detection of potential mosquito breeding sites from aerial images acquired by unmanned aerial vehicles,'' \emph{Computers, Environment and Urban Systems}, vol.~90, p. 101692, 2021.

\bibitem{ml_for_mosquito_control}
A.~Joshi and C.~Miller, ``Review of machine learning techniques for mosquito control in urban environments,'' \emph{Ecological Informatics}, vol.~61, p. 101241, 2021.

\bibitem{zhang2023designing}
K.~ZHANG and C.~C. ZHANG, ``Designing a low-cost microcontroller-based rover for microplastic detection using deep-learning image detection and near-infrared spectroscopy,'' in \emph{2023 Congress in Computer Science, Computer Engineering, \& Applied Computing (CSCE)}.\hskip 1em plus 0.5em minus 0.4em\relax IEEE, 2023, pp. 957--962.

\bibitem{srisuphab2018insect}
A.~Srisuphab, P.~Silapachote, W.~Tantratorn, P.~Krakornkul, and P.~Darote, ``Insect detection on an unmanned ground rover,'' in \emph{TENCON 2018-2018 IEEE Region 10 Conference}.\hskip 1em plus 0.5em minus 0.4em\relax IEEE, 2018, pp. 0954--0959.

\end{thebibliography}

\end{document}